\documentclass[10pt]{article}
\usepackage[utf8]{inputenc}
\usepackage[margin=.8in]{geometry}
\usepackage{rotating}
\usepackage{booktabs}
\usepackage{array}
\usepackage{amsmath, amsfonts,amssymb}
\usepackage{color}
\usepackage{hyperref}
\usepackage{bbm}
\usepackage{algorithm}
\usepackage[noend]{algpseudocode}
\usepackage{subcaption}
\usepackage{graphbox}

\usepackage{mathtools}
\mathtoolsset{showonlyrefs}

\usepackage[
    backend=biber,
    style=ieee,
    citestyle=numeric-comp,
    sorting=none,
    giveninits=true,
    natbib,
    hyperref,
    maxbibnames=99,
    doi=false,isbn=false,url=false,eprint=false
]{biblatex}

\usepackage{amsthm}
\usepackage[normalem]{ulem}
\usepackage{soul}

\newtheorem{prop}{Proposition}

\newtheorem{definition}{Definition}
\newtheorem{remark}{Remark}

\usepackage{chngcntr}
\usepackage{apptools}
\AtAppendix{\counterwithin{theorem}{section}}
\AtAppendix{\counterwithin{prop}{section}}

\usepackage{mathtools}

\usepackage[dvipsnames]{xcolor}

\def\R{\mathbb{R}}
\def\E{\mathbb{E}}

\newcommand{\Y}{\mathcal{Y}}
\newcommand{\X}{\mathcal{X}}

\newcommand{\T}{\mathcal{T}}

\newcommand{\lhat}{\hat{\lambda}}

\newcommand{\ind}[1]{\mathbbm{1}\left\{#1\right\}}

\renewcommand{\P}{\mathbb{P}}

\makeatletter
\def\blfootnote{\xdef\@thefnmark{}\@footnotetext}
\makeatother

\title{Image-to-Image Regression with Distribution-Free Uncertainty Quantification and Applications in Imaging}
\author{Anastasios N. Angelopoulos\footnote{equal contribution}, Amit Kohli$^*$, Stephen Bates,\\
Michael I. Jordan, Jitendra Malik\\ 
\small{
Department of Electrical Engineering and Computer Science, University of California, Berkeley} \\
\\

Thayer Alshaabi, Srigokul Upadhyayula \\
\small{
Advanced Bioimaging Center, Department of Molecular and Cell Biology, University of California, Berkeley} \\
\\
Yaniv Romano \\
\small{
Departments of Electrical and Computer Engineering and of Computer Science, Technion - Israel Institute of Technology}}
\date{\today}

\begin{document}

\maketitle

\begin{abstract}
    Image-to-image regression is an important learning task, used frequently in biological imaging.
    Current algorithms, however, do not generally offer statistical guarantees that protect against a model's mistakes and hallucinations.
    To address this, we develop uncertainty quantification techniques with rigorous statistical guarantees for image-to-image regression problems.
    In particular, we show how to derive uncertainty intervals around each pixel that are guaranteed to contain the true value with a user-specified confidence probability.
    Our methods work in conjunction with any base machine learning model, such as a neural network, and endow it with formal mathematical guarantees—regardless of the true unknown data distribution or choice of model.
    Furthermore, they are simple to implement and computationally inexpensive.
    We evaluate our procedure on three image-to-image regression tasks: quantitative phase microscopy, accelerated magnetic resonance imaging, and super-resolution transmission electron microscopy of a \emph{Drosophila melanogaster} brain. 
\end{abstract}

\section{Introduction}
\label{sec:introduction}
The deployment of image-to-image regression in scientific imaging has generated enormous excitement, promising a future where the resolution of an imaging system can be improved algorithmically~\cite{weigert2018content}.
For example, research developments in machine learning have accelerated MRI scans by an order of magnitude~\cite{zbontar2018fastmri}.
But, to this day, there remains an elephant in the room, obstructing the deployment of these systems: \emph{how can we know when the model has produced an incorrect prediction?}

\begin{figure}[t]
    \centering
    \includegraphics[width=\textwidth]{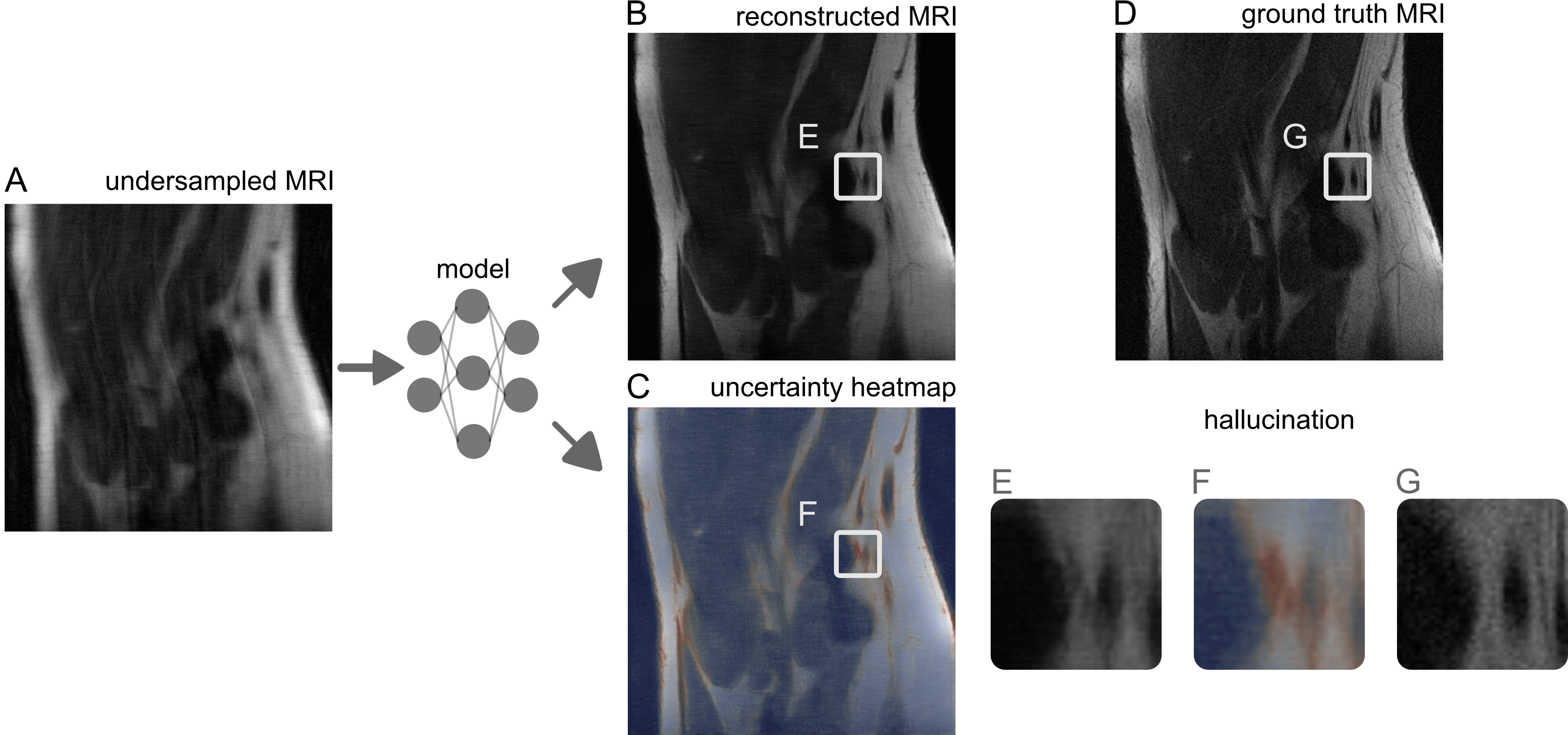}
    \caption{\textbf{An algorithmic MRI reconstruction with uncertainty.} A rapidly acquired but undersampled MR image of a knee (A) is fed into a model that predicts a sharp reconstruction (B) along with a calibrated notion of uncertainty (C). In (C), red means high uncertainty and blue means low uncertainty. Wherever the reconstruction contains hallucinations, the uncertainty is high; see the hallucination in the image patch (E), which has high uncertainty in (F), and does not exist in the ground truth (G). For experimental details, see Section~\ref{subsec:fastmri}.}
    \label{fig:teaser}
\end{figure}

In most cases, we cannot. 
Indeed, the expressive power of modern machine learning is also its torment.
Deep learning models have revolutionized predictive accuracy, but obversely, they fail in silent, unknown, and even unknowable ways.
For scientific imaging settings, where learning will be used for inference and discovery, we need ways to understand when and how a model's predictions might be wrong.
Nonetheless, image-to-image regression algorithms, such as those for denoising and super-resolution, are normally deployed without any notion of statistical reliability.
The scientist is therefore left worrying that their new discovery is simply the model's hallucination.
The purpose of this paper is to introduce a technique which rigorously quantifies the uncertainty in an image-valued point prediction, thereby alerting the scientist of potential hallucinations (see Figure~\ref{fig:teaser}).

We will develop a method for endowing any image-to-image regression model with per-pixel \emph{uncertainty intervals}. 
At a particular pixel, an uncertainty interval is a range of values guaranteed to contain the true value of that pixel with high probability. 
Our contributions are the following:
\begin{enumerate}
    \item We introduce \emph{distribution-free} uncertainty quantification to image-to-image regression; this means the uncertainty intervals will have a rigorous guarantee for any image dataset and any regression model, regardless of the number of data points used to construct the interval. 
    \item We introduce and evaluate several practical algorithms for constructing these sets, including \emph{pixelwise quantile regression}, an extension of quantile regression~\cite{koenker1978regression} to this setting. In experiments, quantile regression consistently leads to the best performance of any uncertainty quantification algorithm, often by a large amount.
    \item We apply our methods to three challenging imaging problems: quantitative phase microscopy, accelerated magnetic resonance imaging (MRI), and super-resolution transmission electron microscopy of a \emph{Drosophila melanogaster} brain.
    \href{https://github.com/aangelopoulos/im2im-uq}{Our accompanying codebase} allows easy application of these methods to any imaging problem, and the exact reproduction of the aforementioned examples.\footnote{\url{https://github.com/aangelopoulos/im2im-uq}}
    The proposed calibrated pixelwise quantile regression approach offers state-of-the-art results on these tasks, in the sense that its uncertainty intervals are smaller than those from other methods.
\end{enumerate}

\subsection{Notation and Goal}
\label{subsec:notation}
The inputs $X$ and outputs $Y$ are both images in $\mathcal{X} = [0,1]^{M \times N}$ (for simplicity of notation, we discuss the case where $X$ and $Y$ are the same size).
We also assume access to an \emph{underlying predictor} $\hat{f}(X)$ mapping from $X$ to a point prediction of $Y$.
The reader can imagine $X$ to be a downsampled version of $Y$, and $\hat{f}(X) \in \R^{M \times N}$ to be a neural network trained to upsample $X$ and reconstruct $Y$ (this is the \emph{super-resolution} task).

Our task is to create uncertainty intervals around each pixel of the predicted image $\hat{f}(X)$ that contain the true pixel values with a user-specified probability. 
Formally, we will construct the following interval-valued function for each pixel,
\begin{equation}
    \label{eq:nested-sets}
    \T(X)_{(m,n)} = \left[\hat{f}(X)_{(m,n)}-\hat{l}(X)_{(m,n)}, \hat{f}(X)_{(m,n)} + \hat{u}(X)_{(m,n)} \right],
\end{equation}
which takes an image and outputs the uncertainty interval for each pixel $(m,n)$.
Notice that the intervals always include the prediction $\hat{f}(X)$, and have width $\hat{l}(X)$
in the lower direction and $\hat{u}(x)$ in the upper direction. 
Intuitively, a large value in $\hat{u}(X)$ indicates a pixel that could have a much higher value than the prediction (undershooting).
Likewise, a large pixel value in $\hat{l}(X)$ indicates a pixel that could have a much lower value than the prediction (overshooting).
We will form the uncertainty intervals by using a held-out set of calibration data, $\{(X_i,Y_i)\}_{i=1}^n$, to assess the model's performance.
The uncertainty intervals will be statistically valid in the following sense. 
The user selects a risk level $\alpha \in (0,1)$, and an error level $\delta \in (0,1)$, such as $\alpha = \delta = 0.1$. 
Then, we construct intervals that contain at least $1-\alpha$ of the ground-truth pixel values with probability $1-\delta$. 
That is, with probability at least $1-\delta$,
\begin{equation*}
    \E\left[\frac{1}{MN}\Big|\big\{(m,n) : Y^{\rm test}_{(m,n)} \in \T(X^{\rm test})_{(m,n)}\big\}\Big|\right] \ge 1 - \alpha,
\end{equation*}
where $X^{\rm test}, Y^{\rm test}$ is a fresh test point from the same distribution as the calibration data. 

In the next section, we will describe in detail the algorithm for generating $\hat{l}(X)$ and $\hat{u}(X)$ as well as its statistical properties. 
Importantly, this algorithm is modular, allowing the user to use the most complex, cutting-edge methods for learning $\hat{f}$ (i.e., the best neural network methods), all the while having uncertainty intervals that reliably communicate the quality of the predictions.

\section{Methods}
\label{sec:methods}

\begin{figure}[t]
    \centering
    \includegraphics[width=0.75\textwidth]{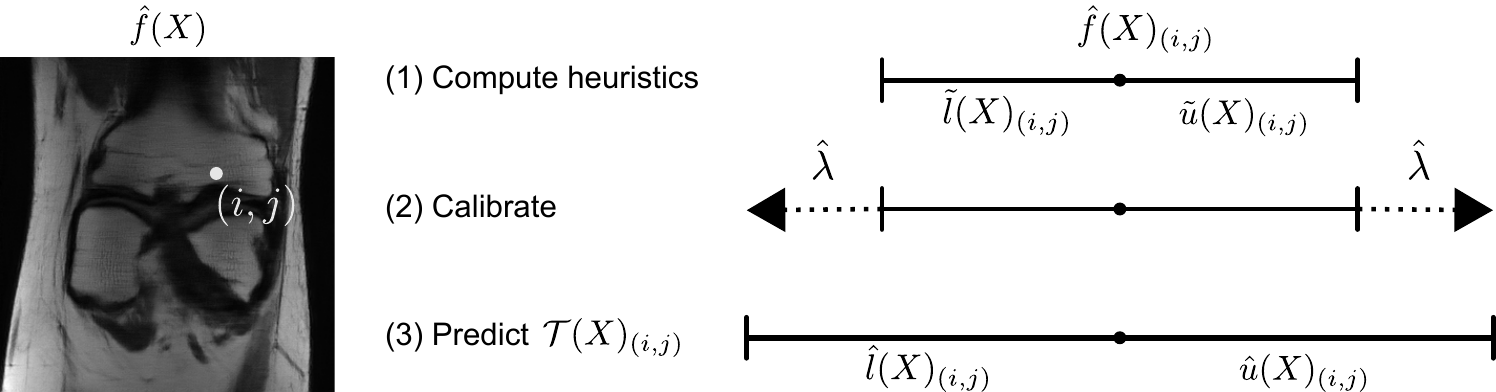}
    \caption{\textbf{An explanation of image-valued risk-controlling prediction sets.} We visualize the process of constructing an uncertainty interval for a single pixel $(i,j)$ of the model's prediction $\hat{f}(X)$. In the first step, we compute the heuristic upper and lower interval lengths. Second, we choose $\hat{\lambda}$ via the RCPS calibration procedure in Section~\ref{subsec:calibration}. Finally, we form the risk-controlling prediction set $\T_{\lhat}(X)_{(i,j)}$ as in~\eqref{eq:nested-sets}.}
    \label{fig:explanatory}
\end{figure}

We now formally describe the method for constructing uncertainty intervals.
Each pixel in the image will get its own uncertainty interval, as in~\eqref{eq:nested-sets}, that is statistically guaranteed to contain the true value with high probability.

The procedure that yields these intervals, visualized in Figure~\ref{fig:explanatory}, has two phases. 
First, we train a model to output a heuristic notion of uncertainty. 
In practice, this amounts to training a machine learning system to output a point prediction $\hat{f}$, a heuristic lower interval length $\tilde{l}$, and a heuristic upper interval length $\tilde{u}$ using any method, such as a neural network.
In Section~\ref{subsec:heuristics} we introduce and benchmark four possible methods of learning these heuristics. 
The uncalibrated intervals $(\hat{f}(X) - \tilde{l}(X), \hat{f}(X) + \tilde{u}(X))$ are heuristic in the sense that they do not contain
the ground truth with the desired probability---we made no assumptions about the algorithm used to train $\tilde{l}$ and $\tilde{u}$. 
To remedy this, in the second phase we calibrate the heuristic notions of uncertainty by scaling them until they contain the right fraction of the ground truth pixels.
That is, we multiply the upper and lower lengths by a value $\hat{\lambda}$ that is chosen using the procedure that we will describe in Section~\ref{subsec:calibration}.  
The final intervals are exactly those in~\eqref{eq:nested-sets}, with the upper and lower widths 
\begin{equation}
    \hat{l}(x) = \hat{\lambda} \tilde{l}(x) \quad \text{ and } \quad \hat{u}(x) = \hat{\lambda} \tilde{u}(x).
\end{equation}
Algorithm~\ref{alg:explanatory} summarizes this process.

Following the above strategy will give us uncertainty intervals that satisfy
the desired statistical guarantee from Section \ref{subsec:notation}.
We call a set of these rigorous uncertainty intervals---one for each pixel in an image---an image-valued \emph{Risk-Controlling Prediction Set}.
\begin{definition}[Risk-Controlling Prediction Set (RCPS), modified from~\cite{bates2021distribution}]
    \label{def:rcps-formal}
    We call a random set-valued function $\T : \X \to \left(2^{[0,1]}\right)^{M \times N}$
    an \emph{($\alpha$,$\delta$)-Risk-Controlling Prediction Set} if 
    \begin{equation}
        \label{eq:rcps-guarantee}
        \P\left(\E\big[L(\T(X),Y)\big] > \alpha \right) \leq \delta,
    \end{equation}
    where 
    \begin{equation}
        L(\T(X),Y) = 1-\frac{\Big|\big\{(m,n) : Y_{(m,n)} \in \T(X)_{(m,n)}\big\}\Big|}{MN}.
    \end{equation}
\end{definition}
\begin{remark}
    The inner expectation in~\eqref{eq:rcps-guarantee} is over a new test point, $(X,Y)$.
    The outer probability is over the calibration data, $\big\{(X_i,Y_i)\big\}_{i=1}^n$.
    In other words, $\T$ is constructed based on the calibration data, which makes it a random function.
    We will only fail to control the risk if we are unlucky with the sample of calibration data, with probability $\delta$.
\end{remark}
Parsing the above equation, we define a level $\alpha$, which tells us what fraction of pixels in the image we allow to fall outside of the intervals. If we set $\alpha=0.1$, for example, it means no more than 10\% of the true pixel values will lie outside of $\T$ except with probability $\delta$.

\begin{algorithm}[t]
\caption{Generating Image-Valued RCPS}
\label{alg:explanatory}

\begin{algorithmic}[1]

\State Train model that outputs point prediction $\hat{f}$ and heuristic lower and upper interval lengths $\tilde{l}$ and $\tilde{u}$.
\State Compute the calibrated parameter $\hat{\lambda}$ using the calibration data and Algorithm~\ref{alg:rcps-cal}.
\State Construct $\T$ as in~\eqref{eq:nested-sets}.
\State For a new image $X$, output the risk-controlling prediction set $\T(X)$.

\end{algorithmic}

\end{algorithm}

Having laid out the goal and general algorithm, we now discuss how to train the model to output heuristic notions of uncertainty for eventual calibration.

\subsection{Picking a Heuristic Notion of Uncertainty}
\label{subsec:heuristics}

The selections of $\tilde{l}$ and $\tilde{u}$ will ultimately determine the properties of the prediction sets, such as their size and shape.
We will learn these heuristics from the same training dataset used to train $\hat{f}$.
Here, we develop four different heuristic notions of uncertainty, which we will evaluate and compare in later experiments (Section~\ref{sec:experiments}).
These heuristics are
\begin{enumerate}
    \item regression to the magnitude of the residual,
    \item parameterizing each pixel as a Gaussian and reporting its standard deviation,
    \item outputting a softmax distribution at each pixel, and
    \item pixelwise quantile regression.
\end{enumerate}

Although each of these methods is trained to predict some form of uncertainty, they may not do it well---hence the need for calibration via Algorithm~\ref{alg:rcps-cal} after training.
Each heuristic requires the use of a different loss function when training the neural network via gradient descent.
The remainder of this subsection describes each loss function precisely.
For notational simplicity, we omit subscripts and sums indexing different pixels; in the experiments, we train our models by averaging the loss function applied to each pixel separately.

\subsubsection{Magnitude of the Residual}
\label{subsec:res-mag}
In this flavor of uncertainty quantification, we set $\tilde{u}=\tilde{l}$, referring to both the upper and lower interval lengths as $\tilde{u}$ (the letter `u' is a mnemonic for the `uncertainty' of the model).
We then optimize $\tilde{u}$ for the following loss function:
\begin{equation}
    \mathcal{L}(x,y) = \left(\tilde{u}(x) - \big|\hat{f}(x) - y\big|\right)^2.
\end{equation}
The loss function encourages each pixel of $\tilde{u}$ to be equal to the model's error at that pixel.
Notice that $\mathcal{L}(x,y)=0$ in the ideal case when the heuristic is exactly equal to the magnitude of the residual, i.e., $\tilde{u}(x) = |\hat{f}(x)-y|$.
Estimating the magnitude of the residual is a straightforward way of quantifying a model's error, although it has two downsides.
Firstly, it can only construct symmetric intervals, which makes the pixelwise intervals less informative and can inflate the set size.
Second, unlike quantile regression, there is no guarantee that the residual estimate results in a valid prediction set without RCPS.
Third, estimating the residual's magnitude is challenging since the training residuals are likely to be smaller than the test ones due to overfitting, unless an extra data split is used.

\subsubsection{One Gaussian Per Pixel}
We will now explain another common heuristic, which involves modeling each pixel as a sample from a Gaussian distribution with a particular mean and standard deviation~\cite{nix1994estimating}.
Translating into our notation, $\hat{f}$ will be the mean function, and $\tilde{u}=\tilde{l}$ will be the standard deviation.
We proceed by minimizing the negative log-likelihood of the Gaussian distribution,
\begin{equation}
    \mathcal{L}(x,y) = \log\big(\tilde{u}(x)\big) + \frac{\big(\hat{f}(x)-y\big)^2}{\tilde{u}(x)}.
\end{equation}

Like the residual magnitude method from Section~\ref{subsec:res-mag}, this heuristic is only suited to symmetric intervals and provides no guarantees of coverage without strong assumptions.
Additionally, unlike the residual magnitude and quantile regression methods, one cannot use data splitting to avoid overconfidence due to overfitting.

\subsubsection{Softmax Outputs}
This next heuristic is most common in classification; indeed, it involves reframing image-to-image regression as a classification problem over a discrete set of pixel values.
The procedure is different from the last two examples; the functions $\tilde{u}$ and $\tilde{l}$ are not equal, and they are not learned directly. 
Instead, we train the network to produce an entire probability distribution, and directly extract all three of $\hat{f}$, $\tilde{u}$, and $\tilde{l}$.

Let us first discretize the possible pixel values into $K$ categories: $\{0, \frac{1}{K-1}, ..., \frac{K-1}{K-1}\}$.
We then associate a discrete label with an otherwise continuous label via the function 
\begin{equation}
    D(y) = \bigg|\left\{ i : i \in 0,1,...,K-1 \text{ and } \frac{i}{K-1} \geq y \right\}\bigg|.
\end{equation}
Intuitively, the function $D(y)$ discretizes $[0,1]$, then bins the pixel accordingly.

This allows us to train the neural network to output distributions over pixel values $\hat{\pi}_y(x)$ estimating the conditional probabilities $\P\left[ Y=y \mid X=x \right]$ via the cross-entropy loss,
\begin{equation}
    \mathcal{L}(x,y) = \frac{1}{MN}\sum\limits_{\substack{1 \leq i \leq M \\ 1 \leq j \leq N}}-\hat{\pi}_{D(y)}(x) + \log \left( \sum\limits_{k=1}^K\exp\big(\hat{\pi}_{D(k)}(x)\big) \right).
\end{equation}
Finally, we can extract the prediction and heuristic uncertainties,
\begin{align}
    \hat{f}(x) &= \frac{1}{K-1} \underset{k}{\arg\max}\; \hat{\pi}_k(x) ; \\
    \tilde{u}(x) &= \frac{1}{K-1}  \mathrm{Quantile}\left(1-\frac{\alpha}{2}, \hat{\pi}(x) \right); \\
    \tilde{l}(x) &= \frac{1}{K-1}  \mathrm{Quantile}\left(\frac{\alpha}{2}, \hat{\pi}(x) \right), \\
\end{align}
where
\begin{equation}
    \mathrm{Quantile}\left(\beta, \hat{\pi}(x) \right) = \min\left\{ K' : \sum\limits_{k=1}^{K'}\hat{\pi}_{k}(x) \geq \beta \right\}.
\end{equation}

The softmax approach requires discretizing $\Y$ into $K$ bins, which can severely limit its utility.
The heuristic can only create prediction sets whose endpoints are multiples of $1/K$, which may make it too conservative.
Furthermore, the output representation can be enormous, making the memory constraints infeasible for large images (e.g., for K=256, the model produces an output of size $M\times N \times 256$).

\subsubsection{Pixelwise Quantile Regression}
This final heuristic is a multi-dimensional version of conformalized quantile regression~\cite{romano2019conformalized,koenker1978regression}.
If we want a 90\% uncertainty interval, then reporting the interval between the estimated 95\% and 5\% quantiles for each pixel is a valid approach.
Thus, we set $\tilde{u}$ to be an estimate of the $1-\alpha/2$ conditional quantile and $\tilde{l}$ to be an estimate of the $\alpha/2$ conditional quantile.
We estimate these pixelwise quantiles with a special loss function called a \emph{quantile loss} (sometimes informally referred to as a \emph{pinball loss}), shown below in its general form for the $\alpha$ quantile and its quantile estimator $\hat{q}_{\alpha}(x)$,
\begin{figure}[H]
    \vspace{-0.5cm}
    \begin{minipage}{0.7\linewidth}
    \begin{equation}
        \mathcal{L}_{\alpha}\big(\hat{q}_{\alpha}(x),y\big) = \big(y-\hat{q}_{\alpha}(x)\big)\alpha\ind{y > \hat{q}_{\alpha}(x)} + \big(\hat{q}_{\alpha}(x) - y\big)(1-\alpha)\ind{y \leq \hat{q}_{\alpha}(x)}.\;\;
    \end{equation}
    \end{minipage}
    \begin{minipage}{0.25\linewidth}
        \centering
        \includegraphics[width=\linewidth]{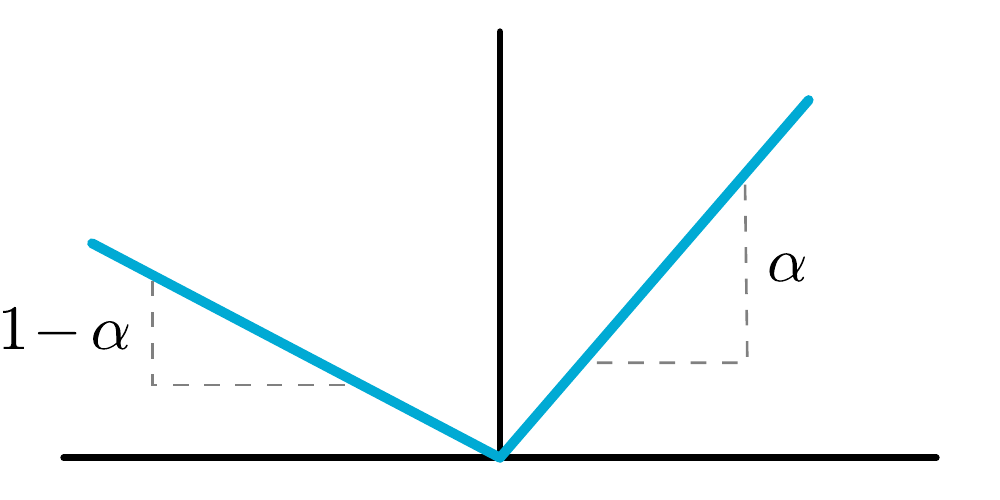}
    \end{minipage}
    \vspace{-0.5cm}
\end{figure}
Omitting some algebra, we can see that the minimizer of this loss is the conditional quantile, i.e., $\mathrm{Quantile}_{Y|X}(\alpha)=\min\{ q : \P\left[ Y < q  \mid X\right] \leq \alpha \}$.
Estimating $\hat{q}$ via empirical risk minimization should therefore approximate the conditional quantile.
This can be made rigorous---under some regularity conditions, quantile regression converges asymptotically to the conditional quantile~\cite{koenker1978regression,chaudhuri1991global,steinwart2011estimating,takeuchi2006nonparametric, zhou1996direct,zhou1998statistical}.
This analysis suggests that quantile regression could be practically effective.

Note that in this case, $\tilde{u}$ and $\tilde{l}$ must be trained with different loss functions, since they estimate different quantiles.
Ultimately, we collapse these into one global loss for the heuristic,
\begin{equation}
    \mathcal{L}(x,y)=\mathcal{L}_{\alpha/2}\big(\tilde{l}(x),y\big) + \mathcal{L}_{1-\alpha/2}\big(\tilde{u}(x),y\big).
\end{equation}
After training, we expect $\tilde{l}$ and $\tilde{u}$ to approximate the $\alpha/2$ and $1-\alpha/2$ quantiles respectively.

\subsection{Calibrating Heuristic Notions of Uncertainty}
\label{subsec:calibration}

\begin{algorithm}[b]
\caption{Pseudocode for computing $\hat{\lambda}$}
\label{alg:rcps-cal}

\textbf{Input:} Calibration data, $(X_i,Y_i)$, $i = 1, \ldots, n$; risk level $\alpha$; error rate $\delta$; underlying predictor $\hat{f}$; heuristic lower and upper interval lengths $\tilde{l}$ and $\tilde{u}$; maximum value $\lambda_{\rm max}$; step size $d\lambda > 0$.

\textbf{Output:} Parameter $\hat{\lambda}$ for computing RCPS.

\begin{algorithmic}[1]

\State $\lambda \gets \lambda_{\rm max}$
\State $r \gets 1$

\While{$r \leq \alpha$}
    \State $\lambda \gets \lambda - d\lambda$
    \For{$i=1,...,n$}  
        \State $L_i \gets L\big(\mathcal{T}_{\lambda}(X_i),Y_i\big)$
    \EndFor
    \State $r \gets \frac{1}{n}\sum\limits_{i=1}^n L_i + \sqrt{\frac{1}{2n}\log \frac{1}{\delta}}$ \Comment{Can replace with any valid upper-confidence bound on the risk.}
\EndWhile

\State $\hat{\lambda} \gets \lambda + d\lambda$ \Comment{Backtrack by one because we overshot.}
\end{algorithmic}

\end{algorithm}

As earlier discussed, we seek to form the RCPS in~\eqref{eq:nested-sets}, which we can compute using any of the heuristics from Section~\ref{subsec:heuristics}.
The function $\T$ will vary based on the heuristic notion of uncertainty used; however, the algorithm for selecting $\lhat$ will provide the guarantee in Definition~\ref{def:rcps-formal} regardless.

The calibration algorithm upper bounds the fraction of pixels falling outside the intervals, and then picks the smallest uncertainty intervals where the upper bound falls below $\alpha$.
Making this more concrete, we index the size of the intervals with a free multiplicative factor $\lambda$,
\begin{equation}
    \T_{\lambda}(X)_{(m,n)} = \left[\hat{f}(X)_{(m,n)}-\lambda \tilde{l}(X)_{(m,n)}, \hat{f}(X)_{(m,n)} + \lambda \tilde{u}(X)_{(m,n)} \right].
\end{equation}
For a particular input image, when $\lambda$ grows, the intervals grow; for a sufficiently large $\lambda$, the intervals will contain all of the ground truth pixel values.
Our job is to pick $\lhat$ to be the smallest value such that $\T_{\lhat}(X)$ satisfies Definition~\ref{def:rcps-formal} (note that $\T(X)=\T_{\lhat}(X)$).
Using the calibration dataset, we form Hoeffding's upper-confidence bound,
\begin{equation}
    \widehat{R}^+(\lambda)= \frac{1}{n}\sum\limits_{i=1}^nL(\T_{\lambda}(X_i),Y_i)+\sqrt{\frac{1}{2n}\log\frac{1}{\delta}}.
\end{equation}
It is shown in~\cite{hoeffding1963} that the Hoeffding bound is valid, that is, $\P\left[\widehat{R}^+(\lambda) < R(\lambda) \right] < \delta$. 
Knowing this, we can use $\widehat{R}^+(\lambda)$ to pick the smallest $\lambda$ satisfying Definition~\ref{def:rcps-formal}.
There is a closed-form expression for this process,
\begin{equation}
    \label{eq:lhat}
    \lhat = \min\left\{\lambda : \widehat{R}^+(\lambda') \leq \alpha, \;\; \forall \alpha' \geq \alpha \right\}.
\end{equation}
\begin{prop}[$\T_{\lhat}$ is an RCPS~\citep{bates2021distribution}]
    \label{prop:rcps-guarantee}
    With $\lhat$ selected as in~\eqref{eq:lhat}, $\T_{\lhat}$ satisfies Definition~\ref{def:rcps-formal}.
\end{prop}
For the proof of this fact, along with a discussion of the tighter confidence bounds used in the experiments, see previous work~\cite{bates2021distribution,angelopoulos2021learn}.
This calibration procedure is easy to implement in code, and we summarize it in Algorithm~\ref{alg:rcps-cal}.

\section{Experiments}
\label{sec:experiments}

The following sequence of experiments applies our methods to several challenging settings in biological imaging.
The goal of these experiments is twofold.
First, we demonstrate the utility of the procedures in practical experiments.
Second, we evaluate the comparative effectiveness of the different heuristics qualitatively, as well as with a series of quantitative metrics.
We will briefly discuss these metrics before providing the details of each experiment.

\subsection{Evaluation Metrics}
\textbf{Empirical risk}. The first quantity to notice is the risk, which should fall below $\alpha$ with probability $(1 - \delta)$.
This is guaranteed in general by Proposition~\ref{prop:rcps-guarantee}.
For each dataset and heuristic, we make a histogram of the risk over several runs of the RCPS calibration, showing it is indeed controlled at the desired level.

\noindent \textbf{Prediction set size}. If the underlying heuristic notion of uncertainty is poor, then, in order to control the risk, the sets may need to be large.
Generally, such an output is not informative to a practitioner, and all else equal, smaller intervals give more actionable assessments of the regression's quality.
Thus, we report histograms of the interval size for each metric---smaller is better.

\noindent \textbf{Size-stratified risk}. Next, we seek prediction sets that do not systematically make mistakes in difficult parts of the image. 
Our risk control requirement in Definition ~\ref{def:rcps-formal} may be satisfied even if the prediction sets systematically fail to contain the most difficult pixels. 
For example, if $\alpha=0.1$ and $90\%$ of pixels are covered by fixed-width intervals of size $0.01$, then the requirement is satisfied---however, the sets no longer serve as useful notions of uncertainty. 
To investigate such behavior, we evaluate the \emph{size-stratified risk}~\cite{angelopoulos2021gentle}---i.e., we stratify pixels by the quartile of their interval sizes, and report the empirical risk within each of these quartiles.
The desire is to have the risk be at approximately the same level for all strata, i.e., the risk should be as similar as possible between pixels with different set sizes.
In other words, when we make a barplot of the stratified risk, the bars should all be the same height.
Achieving this balance means the algorithm is not over-including easier-to-estimate pixels in order to excuse poor performance on difficult ones.

\noindent \textbf{MSE of point prediction}. Finally, we want to pick a heuristic notion of uncertainty that does not harm the accuracy of the point prediction during the joint training process.
To measure this, we plot the \emph{mean-squared error} (MSE) on the validation set for the point prediction which was jointly trained with each heuristic measure of uncertainty.
A lower mean-squared error means that the joint training of the point prediction and heuristic uncertainty worked nicely, and did not degrade the point prediction.
For certain heuristics, such as the Gaussian and softmax versions, this measure is particularly important because these methods do not directly optimize for the MSE and instead require a different procedure for supervising the point prediction (maximizing the Gaussian log likelihood and minimizing the cross-entropy loss, respectively).

\noindent \textbf{Visualizations}. In addition to the quantitative metrics, there is no substitute for seeing visualizations of the uncertainty intervals.
For each example, we show the input, output, and the prediction sets generated by quantile regression along with the ground truth target.
We represent the prediction sets by passing the pixelwise interval lengths through a colormap, where small sets render a pixel blue and large sets render it red. The interpretation, then, is that the redder a region is, the more uncertain it is and conversely, the bluer a region is, the more confident it is.
Consequently, we expect the colormap to be red where the model is missing biological features and around fine structures such as edges which are difficult to reconstruct from partial data.

\subsection{Experimental Details}
We use a standard experimental pipeline for all of the forthcoming experiments.
In all experiments, we fit the predictor $\hat{f}$ and the heuristic notions of uncertainty $\tilde{u}$ and $\tilde{l}$ jointly.
To ensure a level playing field and to promote reproducibility, the code used to define, train, and evaluate the model is shared among all heuristics and datasets.
In order to run a new experiment (e.g., on a new dataset or with a new heuristic), minimal additional code is needed.
We believe this code to be a primary contribution of this paper with utility to researchers and practitioners alike.
It lives at the following open-sourced GitHub link:~\textcolor{blue}{\url{https://github.com/aangelopoulos/im2im-uq}}.

In each experiment, an 8-layer U-Net~\cite{ronneberger2015u} is used as the base model architecture and trained with an Adam optimizer for 10 epochs. 
We swept over two learning rates, $\{0.001,0.0001\}$, and chose the learning rate that minimized the point prediction's MSE for each method in each experiment.
All images get normalized to the interval $[0,1]$.
For the softmax heuristic, we discretized the prediction space with $K=50$ because larger choices of $K$ become too computationally expensive due to the amount of memory needed to store the extra dimension---a major practical limitation of this heuristic.
We choose $\alpha=\delta=0.1$ for the RCPS procedure in all cases, and adaptively select a grid of 1000 values of $\lambda$ for each experiment.
We evaluate each method by plotting its risk, average set size, size-stratified risk, and mean-squared error of the jointly trained prediction, as well as displaying an example. 
Further experimental details are available in the codebase.

We now discuss each experiment in detail.
For each experiment, we include a brief background of the imaging problem, a description of the inputs $X$ and outputs $Y$, and the aforementioned evaluation metrics for each heuristic.

\subsection{Quantitative Phase Microscopy of Leukocytes}
\label{sec:qpi}

\begin{figure}[h]
    \centering
    \includegraphics[width=0.24\linewidth]{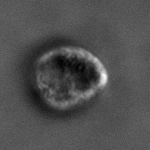}
    \includegraphics[width=0.24\linewidth]{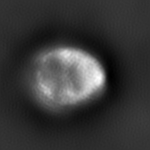}
    \includegraphics[width=0.24\linewidth]{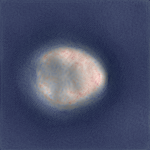}
    \includegraphics[width=0.24\linewidth]{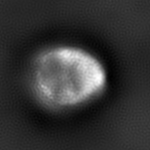}
    \caption{\textbf{Examples of quantitative phase reconstructions with uncertainty} shown in the following order: input, prediction, uncertainty visualization, ground truth. We use the quantile regression version of the procedure.}
    \label{fig:bsbcm-examples}
\end{figure}

\begin{figure}[t]
    \includegraphics[width=0.32\linewidth]{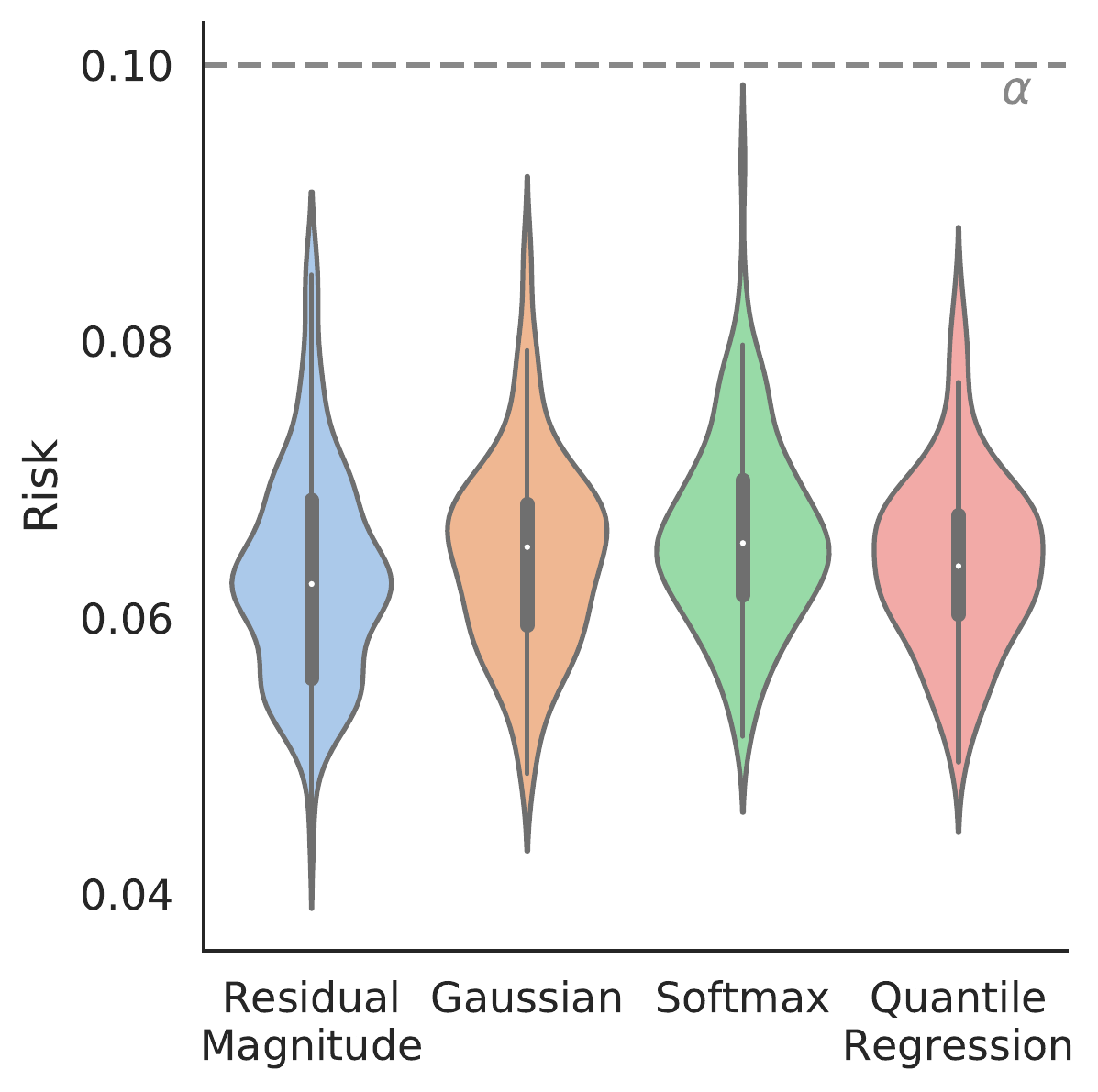}
    \includegraphics[width=0.32\linewidth]{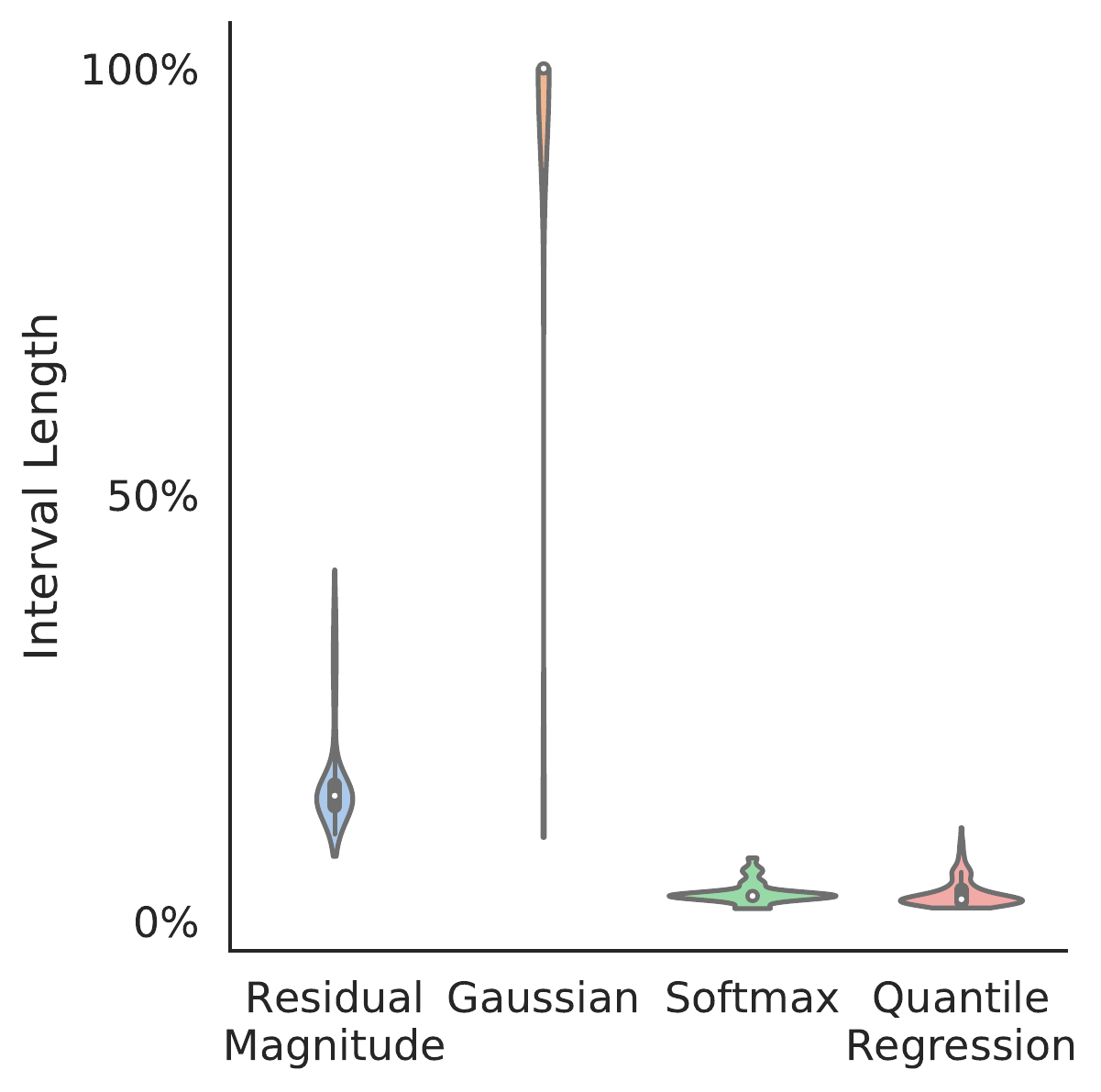}
    \includegraphics[width=0.32\linewidth]{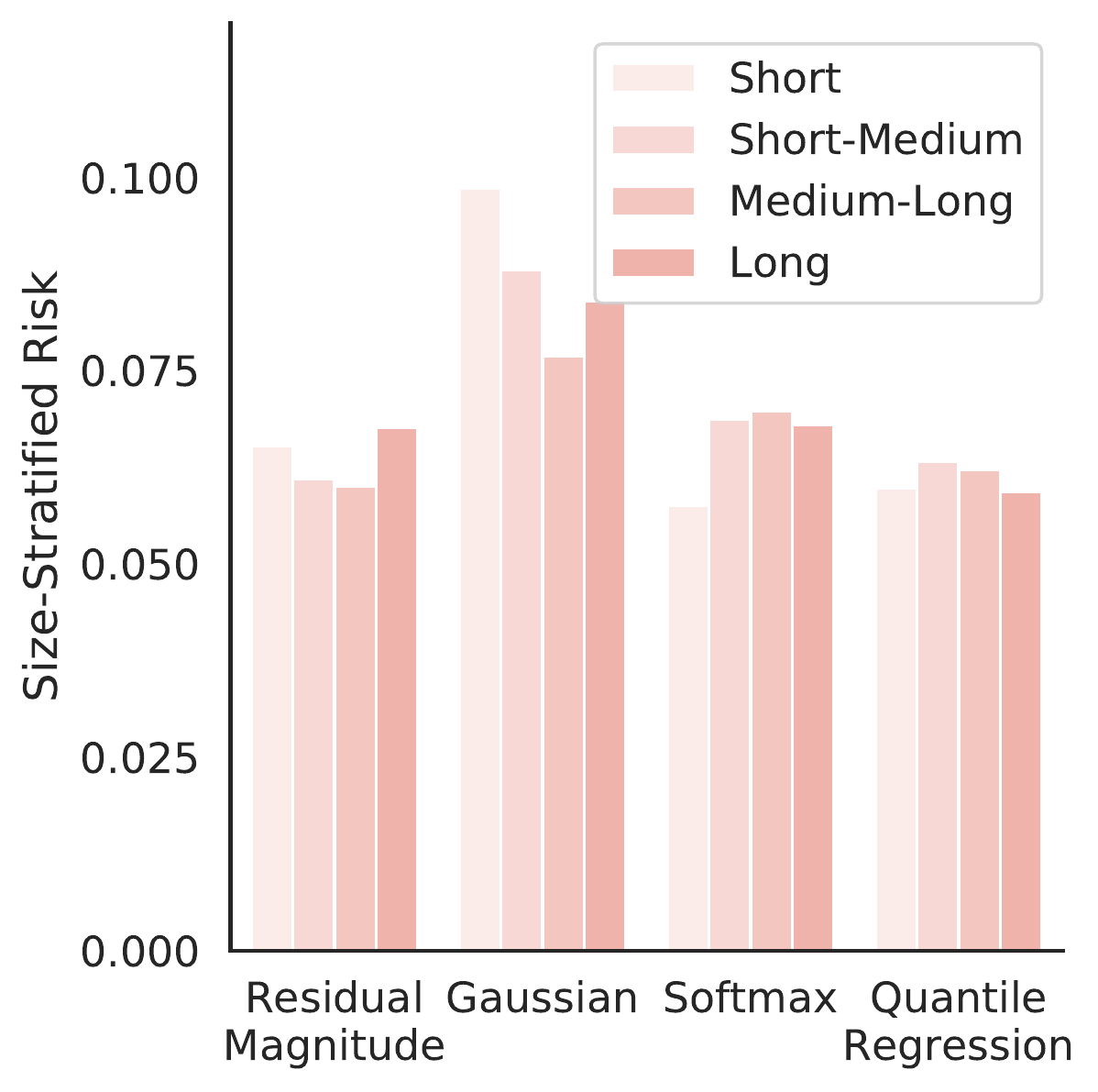}
    \includegraphics[width=0.7\linewidth]{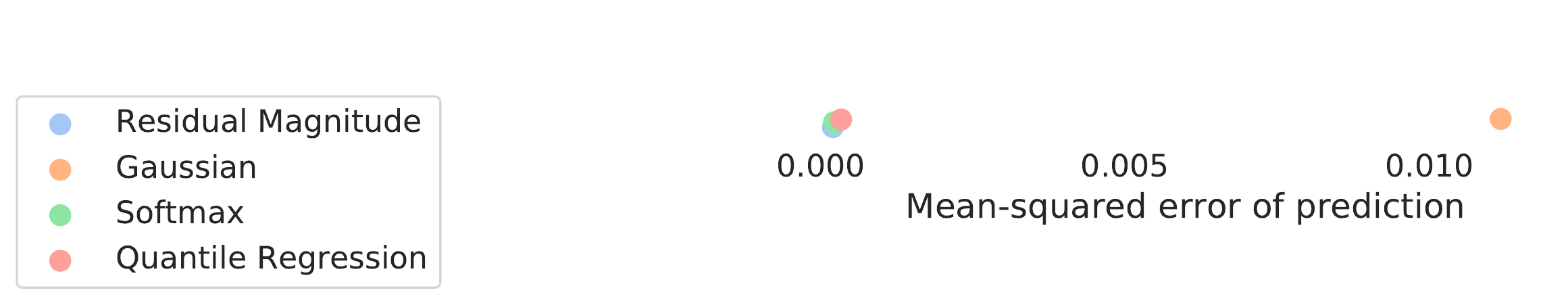}
    \caption{\textbf{A quantitative summary of all four heuristics} after RCPS calibration in the quantitative phase example. All methods control the risk, and quantile regression has the smallest set size. The Gaussian method has poor MSE, interval size, and size-stratified risk because it did not converge in training for either of the learning rates we chose.}
    \label{fig:bsbcm-quantitative}
\end{figure}
\noindent\textbf{Background.} In order to image the structure of cells---which are essentially transparent bags of water---one must measure their local refractive index, or equivalently, the phase delay incurred by light passing through each region of the cell.
This task, known as quantitative phase imaging or QPI~\cite{mir2012quantitative,park2018quantitative,jo2018quantitative}, requires an algorithm to map intensity-only images to the phase value at each pixel, since it is impossible to directly measure the phase of light.
Generally, as input to the algorithm, these methods take in a diverse set of intensity images captured under different imaging conditions, such as the illumination angle. Their performance improves with more input images. 

\noindent\textbf{Dataset description.} Framing QPI as image-to-image regression, we take the input $X$ to be the concatenation of two obliquely illuminated cell intensity images (from opposite angles) and the target $Y$ to be a reference phase image. 
$Y$ is obtained using an analytic phase recovery technique known as differential phase contrast (DPC), which takes four or more images as input~\cite{tian2015quantitative}.
The utility, then, of the regression model is to reduce the number of input images needed for high quality phase prediction, thereby improving the tradeoff between acquisition speed and prediction accuracy.
Furthermore, the quantitative phase values have an intrinsic meaning, so by adding uncertainty intervals which are on the scale of the phase values, we provide an important inferential tool for analyzing cell morphology.

For the experiment, we use the Berkeley Single Cell Computational Microscopy (BSCCM) dataset~\cite{pinkard}, which contains 2,000 single leukocyte (white blood cell) images with 150x150 pixels taken using several imaging modalities. 
Of particular interest to us, this dataset includes images taken under a variety of different angles of illumination co-registered with quantitative phase maps obtained via 4-image DPC.
As input to the U-Net, we concatenate two obliquely illuminated cell images along the channel dimension.
We use 1800 randomly selected data points with a batch size of 64 to train the model, 100 points for calibration, and 100 points for validation.
Our results are visualized in Figure~\ref{fig:bsbcm-examples}.

\noindent\textbf{Results.} We report our results in Figure~\ref{fig:fastmri-quantitative}.
As promised by the calibration procedure, risk-control holds for all choices of heuristic uncertainties. 
In terms of statistical power, we see that quantile regression outcompetes the other heuristics in the trifecta of evaluations---it has the smallest average set size and best size-stratified coverage while remaining competitive with other methods in mean-squared error.
Altogether, these metrics express that the uncertainty intervals produced by calibrated quantile regression are tight and adaptive to the model's performance, even among different pixels within a single prediction.
The softmax heuristic, though seemingly competitive in these evaluations, gives nearly fixed-width intervals, most of which have exactly the same size because of the discretization.

\subsection{Fast Magnetic Resonance Imaging}
\label{subsec:fastmri}
\noindent\textbf{Background.} Much like our previous example, in MRI there exists a tradeoff between imaging speed and quality. 
MRI directly samples an object's spatial frequency (k-space) over time; so it is possible to reduce the scan time by lowering the effective sampling rate in k-space.
Although fast imaging is more comfortable for human subjects and also critical for certain fast movements like the beating of the heart, insufficient sampling results in low quality, aliased MR images. 
However, with deep learning, we can try to fill in the information lost by undersampling to emulate fully sampled images, thereby getting the joint benefits of fast scan times and high quality reconstructions.

\noindent\textbf{Dataset description.} The inputs $X$ are the undersampled images formed by downsampling k-space by a factor of four along a single dimension (the phase encoding direction), and then taking an inverse Fourier transform.
Our outputs $Y$ are the fully sampled MR images. Successfully regressing $X$ to $Y$ essentially accelerates the MRI scan time by a factor of four.

We use the FastMRI dataset for this example~\cite{zbontar2018fastmri}.
The dataset includes 10,000 clinical knee MR volumes taken with 3T or 1.5T magnets which are algorithmically undersampled with k-space masks that emulate fast sampling strategies.
We dissect the volumes into 27,993 randomly selected 320x320 pixel coronal knee slices for training the model, 3,474 for the RCPS calibration, and 3,474 for validation.
We use a batch size of 78.
For the Gaussian method, we standardized the output space to be mean zero and unit variance, since it failed to properly train when normalized to fall in the interval $[0,1]$.

\begin{figure}[ht]
    \centering
    \includegraphics[width=0.32\linewidth]{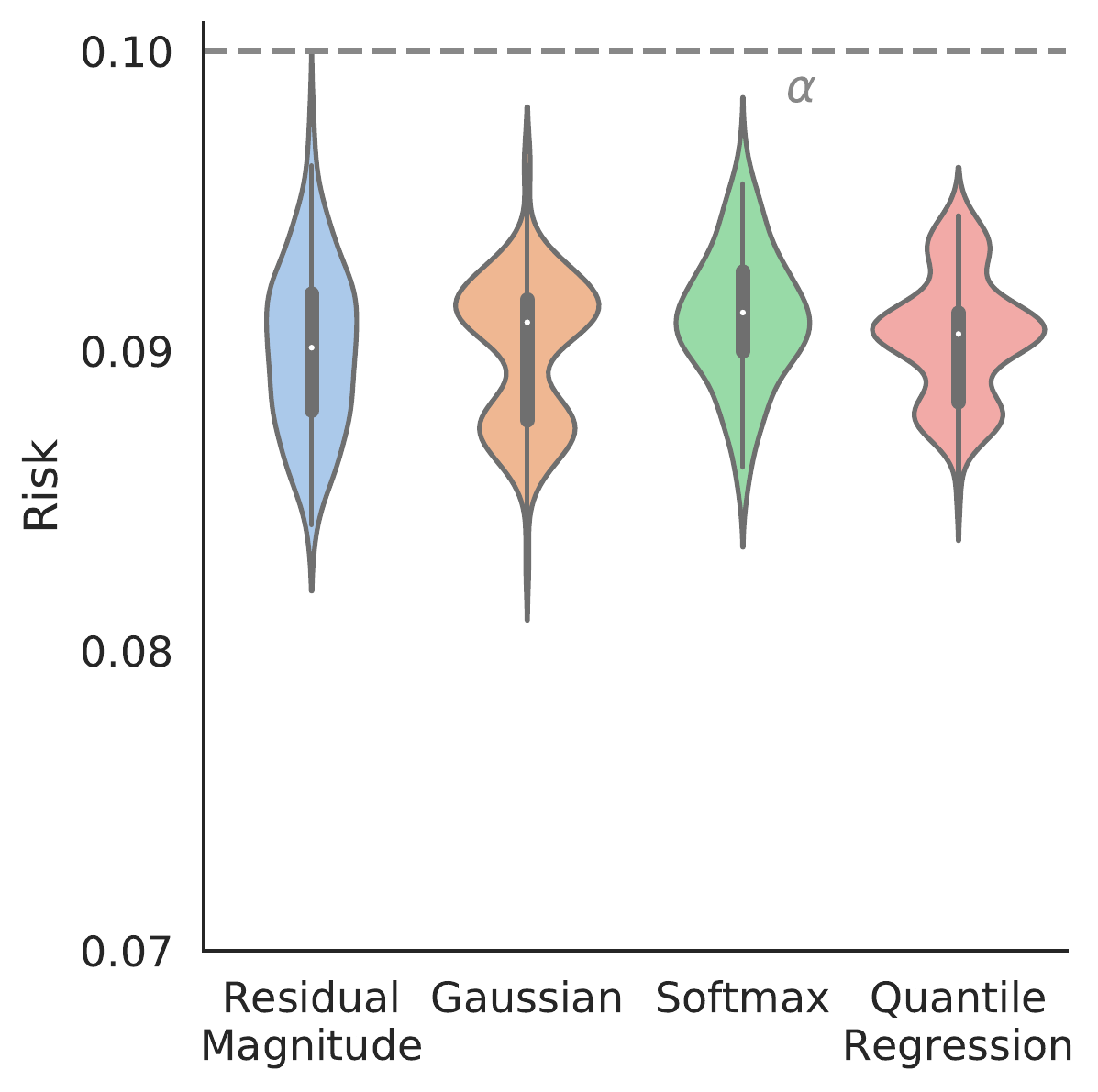}
    \includegraphics[width=0.32\linewidth]{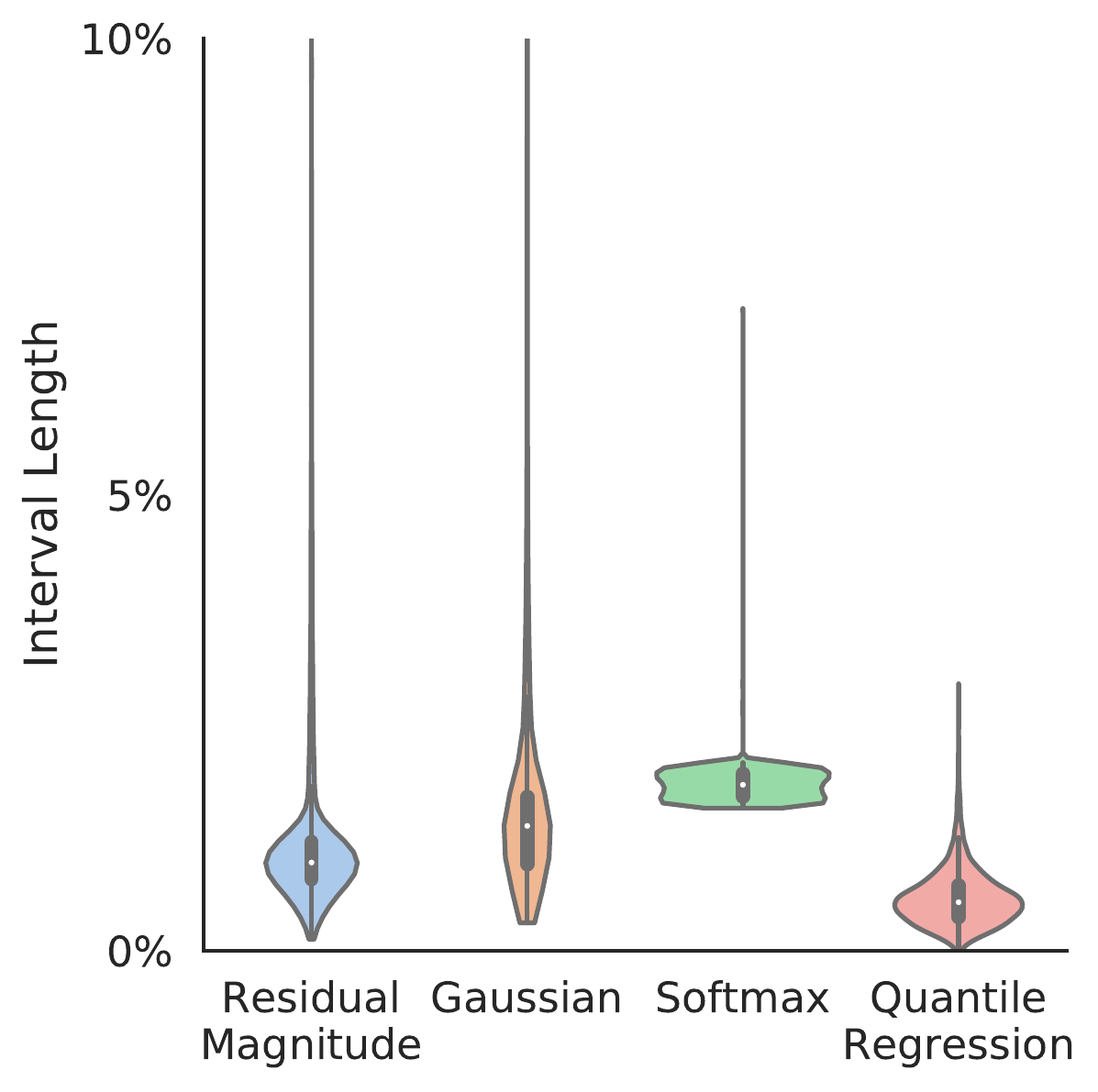}
    \includegraphics[width=0.32\linewidth]{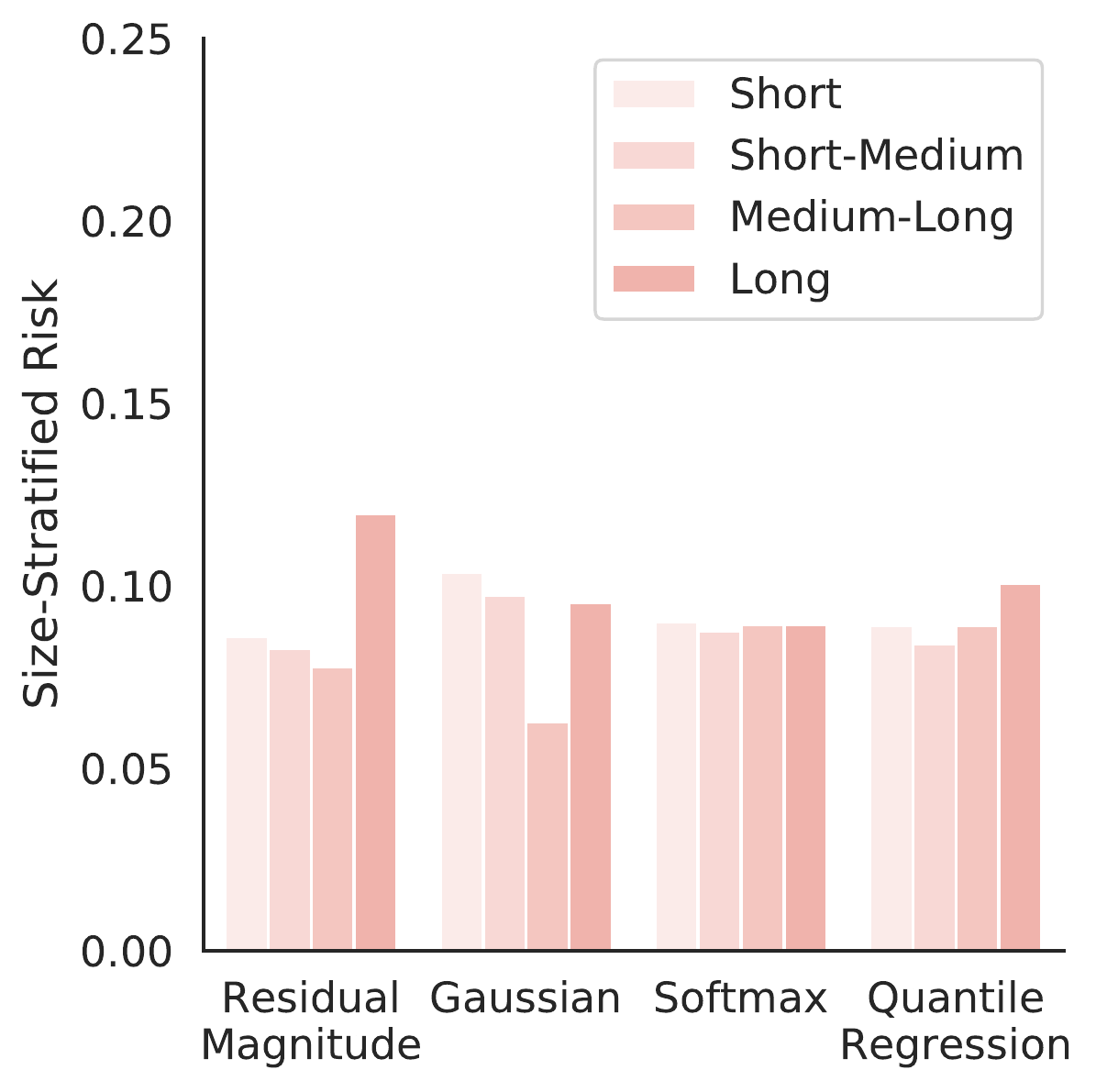}
    \includegraphics[width=0.7\linewidth]{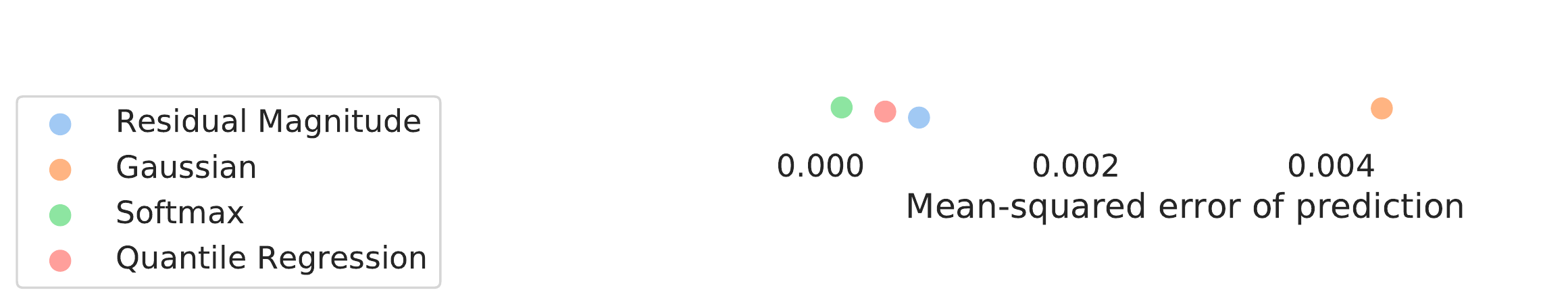}
    \caption{\textbf{A quantitative summary of all four heuristics} after RCPS calibration in the FastMRI example. All three methods control the risk. Quantile regression has the smallest interval size and best size-stratified risk.}
    \label{fig:fastmri-quantitative}
\end{figure}

\noindent\textbf{Results.} Qualitatively, Figure~\ref{fig:teaser} shows an example of an MRI reconstruction using calibrated quantile regression. 
The predictions are slightly blurred versions of the ground truth, likely due to the network's bias toward low frequency outputs~\cite{pmlr-v97-rahaman19a}. The uncertainty intervals have large values in areas with high contrast, expressing the intrinsic uncertainty in localizing edges using incomplete information.
The quantitative results of this experiment, visualized in Figure~\ref{fig:fastmri-quantitative}, are in line with those of the QPI experiment; we achieve the desired risk level and quantile regression performs best on all metrics.
Although the softmax heuristic has near-even size-stratified risk, this is because it outputs quantized sets of nearly fixed size, and the strata are therefore decided by random tie-breaking.

\subsection{Drosophila Brain Transmission Electron Microscopy}

\begin{figure}[h]
    \centering
    \includegraphics[width=0.24\linewidth]{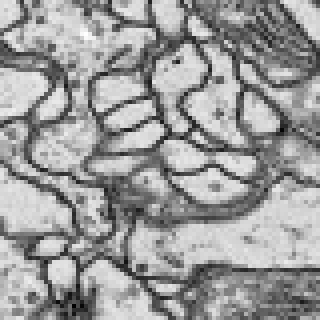}
    \includegraphics[width=0.24\linewidth]{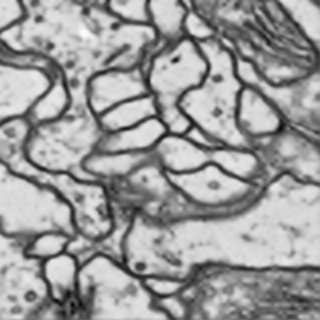}
    \includegraphics[width=0.24\linewidth]{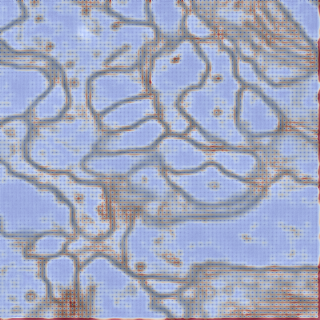}
    \includegraphics[width=0.24\linewidth]{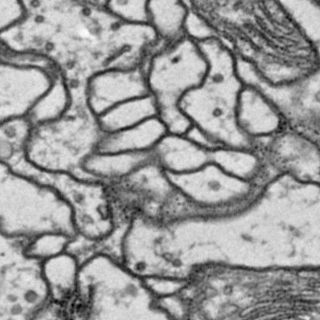}
    \caption{\textbf{Examples of Drosophila brain reconstructions with uncertainty} shown in the following order: input, prediction, uncertainty visualization, ground truth. We use the pixelwise quantile regression version of the procedure.}
    \label{fig:flybrain-examples}
\end{figure}
\begin{figure}[t]
    \includegraphics[width=0.32\linewidth]{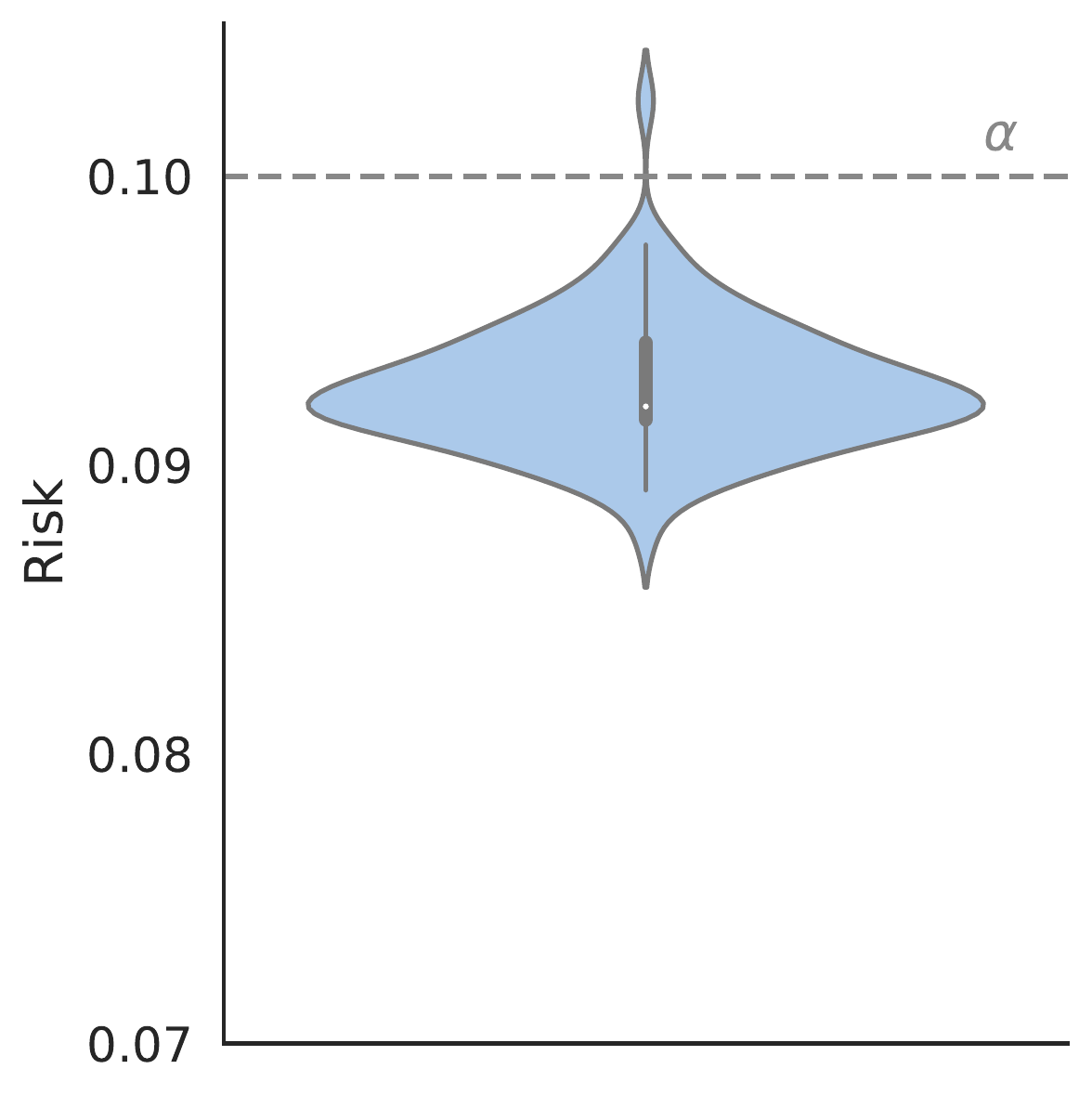}
    \includegraphics[width=0.32\linewidth]{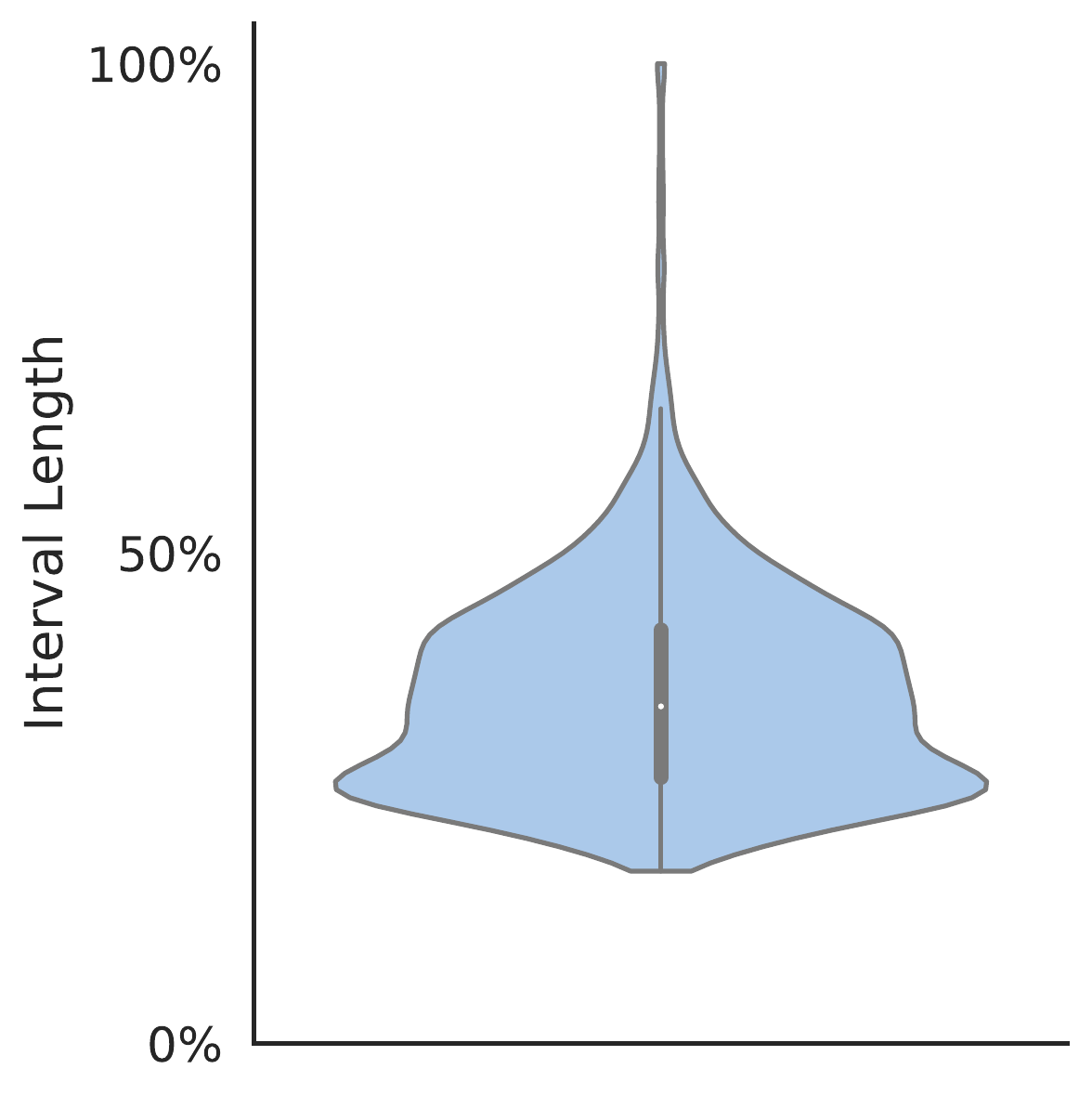}
    \includegraphics[width=0.32\linewidth]{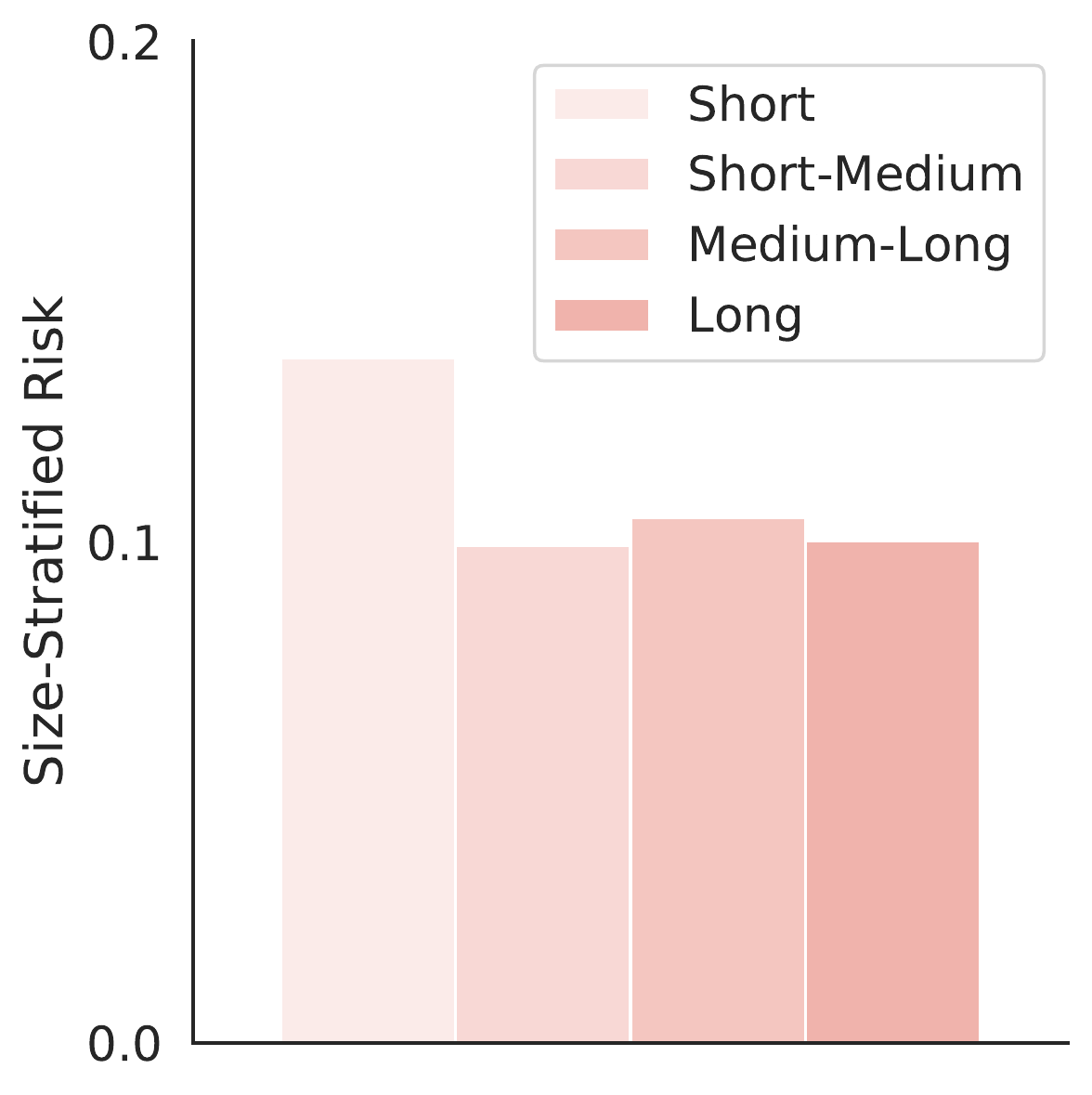}
    \caption{\textbf{Quantitative results of pixelwise quantile regression} on the TEMCA fly brain dataset. The risk is controlled, the intervals have reasonable lengths, and the size-stratified risk is roughly balanced, although slightly more permissive with small intervals. }
    \label{fig:flybrain-quantitative}
\end{figure}

\noindent\textbf{Background.} Finally, we perform algorithmic super-resolution transmission electron microscopy (TEM) of the brain of a Drosophila melanogaster (fruit fly).
TEM uses focused electron beams rather than visible light to produce images, and due to the small de Broglie wavelength of electrons, it can achieve significantly higher resolution than visible light microscopy, on the order of single nanometers.
However, TEM sequentially scans over the sample volume, imaging point-by-point; thus, its scan time scales cubically with the desired resolution. For large volumes like the fly brain shown in Figure~\ref{fig:flybrain-examples}, imaging can take years.
Upsampling lower resolution (say 16nm) TEM data to high resolution (4nm) images could therefore save months of time.

\noindent\textbf{Dataset description.} We consider super-resolution as image-to-image regression, where $X$ is a manually downsampled version of a 4nm TEM image $Y$. In particular, we consider a factor of 4x nearest-neighbor downsampling along both image dimensions to emulate the acquisition of a 16nm TEM image.

We use the Janelia Transmission Electron Microscopy Camera Array (TEMCA2) dataset of the Full Adult Fly Brain~\cite{zheng2018complete}. This dataset contains a 26 TB fly brain volume at four nanometer resolution isotropically along all dimensions. As a consequence of this data burden, we did not run the full suite of procedures on the dataset, and instead ran only the consistently best performing method---pixelwise quantile regression.
The dataset cannot be stored normally and must be chunked \emph{on-the-fly} into patches, so exact dataset sizes cannot be known in advance. We used roughly 2M images of size 320x320 for training, 25K images for calibration, and 25K images for validation.
We use a batch size of 16.
Each image is only seen once.

\noindent\textbf{Results.} The results of this procedure are depicted in Figure~\ref{fig:flybrain-examples}.
The quantitative measures in Figure~\ref{fig:flybrain-quantitative} are similar to those of past experiments. 
Qualitatively, the prediction sets identify regions of high contrast as more uncertain, perhaps due to the spectral bias of CNNs.
The sets are periodically zero-length, i.e., fully confident, every four pixels. 
This is because the input is a downsampled version of the target, so the model perfectly knows every fourth pixel.
Consequently, at those pixels, the model does not have any uncertainty. 
This highlights the adaptivity and tightness of the prediction sets; they are not only useful in understanding where the model is poor, but also where the model performs reliably.

\section{Discussion}
The experimental findings indicate that pixelwise quantile regression has strong performance while avoiding many of the drawbacks of other methods.
The softmax method is prohibitively memory-intensive, while achieving slightly worse results than quantile regression in Section~\ref{sec:qpi}.
The Gaussian method, all else equal, seems more difficult to train, and the \texttt{GaussianNLL} loss adversely affects both the prediction quality and the heuristic.
The two winners seem to be regression to the magnitude of the residual and pixelwise quantile regression.
Ultimately, we suggest pixelwise quantile regression because it supports asymmetric intervals, while also achieving better empirical performance.
We hope our sequence of large-scale imaging examples indicate the broad utility of the techniques.

\section{Related Work}
\textbf{Image-to-image regression.} The problem of image-to-image regression has existed for several decades under a variety of names.
The most fundamental class of image-to-image regression problems involves interpolating between samples of a digital image, a setting currently referred to as \emph{image super-resolution}.
Since this problem simplifies to interpolating between discrete samples of a two-dimensional function, methods such as linear interpolation, used as early as 200BC in \emph{The Nine Chapters on the Mathematical Art}~\cite{nineChapters200BC,needham1959science} and 200AD in Ptolemy's \emph{Almagest}~\cite{ptolemy200ADalmagest}, remain commonly deployed today.
Similarly, over a hundred years of research on optimal interpolation, such as that on spline approximations~\cite{hahn1918interpolationsproblems,whitney1953polya,walsh1962best}, has been used in image super-resolution since the 1980s~\cite{keys1981cubic} and continues to be applied.
In the 21st century, learning-based approaches~\cite{freeman2002example,chang2004super,yang2010image} have dominated the research conversation, particularly those using convolutional neural networks~\cite{dong2014learning,dong2015image,johnson2016perceptual} and generative adversarial networks~\cite{goodfellow2014generative,ledig2017photo}.
Beyond interpolation, image-to-image regression also encompasses denoising~\cite{buades2005review,buades2005nonlocal,burger2012image,tian2020deep,goyal2020image}, style transfer~\cite{gatys2016image,isola2017image,zhu2017unpaired,jing2020style}, image colorization~\cite{zhang2016colorful}, and so on. 
A line of work based on the U-Net~\cite{ronneberger2015u} adapts the above techniques for biomedical imaging problems, achieving strong results~\cite{zbontar2018attention,zhou2018unetplusplus}.
We build directly on this line of work.

\textbf{Heuristic notions of uncertainty.} The idea of assuming the output of a neural network has a Gaussian distribution and maximizing its log-likelihood with gradient descent has been employed since at least 1994~\cite{nix1994estimating}.
Although the idea of estimating the magnitude of the residual with a Gaussian process been suggested~\cite{Qiu2020Quantifying}, we are unaware of any papers that use the exact formulation in Section~\ref{subsec:res-mag}, and this is far from a standard method.
The idea of the cross-entropy loss leading to the softmax distributional estimate has its roots in the Kraft-McMillan theorem~\cite{kraft1949device,mcmillan1956two} and related information-theoretic concepts~\cite{cover1999elements}.
Quantile regression was proposed in the mid-1970s by Koenker and Bassett~\cite{koenker1978regression}.
Since then, many papers have used the technique, applying it to economics~\cite{chaney2011quality,mckenzie2007network,machado2005counterfactual}, machine learning~\cite{hwang2005simple,meinshausen2006quantile,natekin2013gradient}, medical research~\cite{armitage2008statistical}, and more.
A large and vibrant community continues to work today on quantile regression, exploring variations of the technique and their operating characteristics under various conditions, such as for local polynomials~\cite{chaudhuri1991global}, in additive models~\cite{koenker2011additive}, for improved conditional coverage~\cite{feldman2021improving}, and in deep learning~\cite{fort2019deep}.
Unlike the other heuristics discussed, quantile regression comes with an asymptotic guarantee of conditional coverage under certain weak regularity conditions~\cite{koenker1978regression,chaudhuri1991global,zhou1996direct,zhou1998statistical,steinwart2011estimating,takeuchi2006nonparametric}.
Accessible and complete references to the topic of quantile regression are provided in the references~\cite{koenker2005quantile,koenker2018handbook}.
The calibrated, pixelwise version of quantile regression extends these procedures to the image-to-image regression case.
A large literature on Bayesian and approximately Bayesian uncertainty quantification also exists, including MC-Dropout~\cite{gal2016dropout}. 
Ensemble methods~\cite{hansen1990neural,lakshminarayanan2017simple,fort2019deep} are also common for uncertainty quantification in deep learning.
These heuristics, which do not have finite-sample guarantees, fall outside the scope of our discussion, and we refer the reader to~\cite{gal2016bayesian} for an introduction to that area.

\textbf{Distribution-Free Uncertainty Quantification.} Conformal prediction is a general, lightweight procedure for creating uncertainty intervals from any heuristic with finite-sample coverage while requiring no model retraining~\cite{vovk1999machine,vovk2005algorithmic,lei2013conformal,lei2013distribution,lei2014classification,Sadinle2016LeastAS}. 
Of particular interest to us is the method of conformalized quantile regression (CQR)~\cite{romano2019conformalized}.  
We directly build on CQR, by replacing the the conformal subroutine with the fixed-sequence testing procedure from ~\cite{bates2021distribution,angelopoulos2021learn}.
Other works have applied distribution-free uncertainty quantification to biological and medical computer vision tasks~\cite{hechtlinger2018cautious,cauchois2020knowing,romano2020classification,angelopoulos2020sets,angelopoulos2021private,lu2021fair}.
However, we are not aware of any that have studied image-to-image regression.
An introduction to these topics is available in~\cite{angelopoulos2021gentle}.

\printbibliography

\clearpage
\appendix

\end{document}